\documentclass[letterpaper]{article} 
\usepackage{aaai24}  
\usepackage{times}  
\usepackage{helvet}  
\usepackage{courier}  
\usepackage[hyphens]{url}  
\usepackage{graphicx} 
\usepackage{natbib}  
\usepackage{caption} 
\usepackage{algorithm}
\usepackage{algorithmic}
\usepackage{comment}
\usepackage{amsmath}
\usepackage{amsfonts}

\usepackage{soul}

\usepackage{newfloat}
\usepackage{listings}
\DeclareCaptionStyle{ruled}{labelfont=normalfont,labelsep=colon,strut=off} 
\floatstyle{ruled}
\newfloat{listing}{tb}{lst}{}
\floatname{listing}{Listing}
\title{Fairness in Reinforcement Learning: A Survey}
\author {
    Anka Reuel\textsuperscript{\rm 1},
    Devin Ma\textsuperscript{\rm 2}
}
\affiliations {
    \textsuperscript{\rm 1}Stanford University\\
    \textsuperscript{\rm 2}University of Pennsylvania\\
    anka@cs.stanford.edu, 
    devinma@upenn.edu
}

\usepackage{bibentry}

\begin{document}

\maketitle

\begin{abstract}
While our understanding of fairness in machine learning has significantly progressed, our understanding of fairness in reinforcement learning (RL) remains nascent. Most of the attention has been on fairness in one-shot classification tasks; however, real-world, RL-enabled systems (e.g., autonomous vehicles) are much more complicated in that agents operate in dynamic environments over a long period of time. To ensure the responsible development and deployment of these systems, we must better understand fairness in RL. In this paper, we survey the literature to provide the most up-to-date snapshot of the frontiers of fairness in RL. We start by reviewing where fairness considerations can arise in RL, then discuss the various definitions of ``fairness'' in RL that have been put forth thus far. We continue to highlight the methodologies researchers used to implement fairness in single- and multi-agent RL systems before showcasing the distinct application domains that fair RL has been investigated in. Finally, we critically examine gaps in the literature, such as understanding fairness in the context of RLHF, that still need to be addressed in future work to truly operationalize fair RL in real-world systems.
\end{abstract}

\section{Introduction} \label{intro}


As artificial intelligence (AI) has been thrust into the public spotlight by the recent proliferation of generative AI applications, so has the issue of potential harm that AI can do to users and society in general \citep{bommasani2021opportunities}. One particular issue is fairness. Before the advent of machine learning (ML), fairness was largely studied as a philosophical and societal concept \citep{rawls1971atheory, wolff1998fairness, rabin1993incorporating}. Nowadays, it is an active field of research \citep{corbett2018measure, mehrabi2021survey}, studied with mathematical rigor by theorists and practitioners alike. Such efforts have enabled industry to start operationalizing both the implementation of fairness-aware algorithms and auditing fairness in products and services. However, one important area of ML where the study of fairness is still nascent is reinforcement learning (RL), in which autonomous agents interact in an environment to maximize a pre-specified reward \citep{sutton2018reinforcement}. Whereas most efforts so far in fairness focus on one-shot classification and decision-making tasks \citep{mehrabi2021survey}, real-world systems, especially in robotics, often involve autonomous agents (e.g., the robot) interacting with the environment over a longer period of time. In such environments, the agent's decision now will impact its future rewards. Often, these agents also impact humans during their interaction with the environment. Because RL is at the heart of many modern autonomous systems, including robotics, autonomous driving, and healthcare, the study of fairness in RL is particularly relevant today.\\


Typically, RL maximizes the long-term discounted reward of an agent where the reward is tailored to one specific objective \citep{sutton2018reinforcement}. This reward can be, for example, overall output, points in a game, or stock market gains. While this ensures that the reward with respect to the defined objective is maximized, it does not necessarily involve considerations of group or individual fairness \citep{weng2019fairness}. Only focusing on performance rewards but ignoring fairness notions comes with various negative ramifications in certain RL systems. First, using RL algorithms without fairness constraints can result in potentially discriminatory outcomes that are unethical and harmful. Depending on the application the algorithm is used in, unfair outcomes further have the potential to propagate existing societal biases \citep{vlasceanu2022propagation}. Second, in many developed countries, explicit mandates exist to guarantee fairness \citep{hacker2018teaching}. Developers using RL-based systems need to ensure, by law, that their algorithms do not discriminate on protected attributes like gender, race, or religion. Third, if certain groups are underrepresented, user-centric systems might be abandoned altogether by the unfairly treated groups and other users. For example, if an RL algorithm pushes more white content creators on a platform like YouTube than minority creators, the latter might leave the platform due to inadequate exposure. Subsequently, the platform's content may become more homogeneous, leading to less satisfied users \citep{liu2020balancing}. This outcome is undesirable to both the users and the company that markets the product. Since \citet{jabbari2017fairness} initiated the study of fairness in RL in 2017, there has been a growing interest in studying the theory and applications of fair RL. Despite these efforts, our understanding of fundamental questions about fairness in RL remains nascent. This issue necessitates a comprehensive survey of the field to provide a snapshot of the frontiers of fairness in RL, offer a review of the definitions and methods to incorporate fairness in RL, and consider the implications of incorporating fairness into RL-enabled socio-technical systems.\\





 
Earlier work by \citet{zhang2021survey} presented a survey on fairness in learning-based sequential decision-making without considering bandit settings or a critical comparative analysis of definitions, methods, or application areas of fair RL. To our knowledge, only \citet{gajane2022survey} has attempted to survey the field as we do. While they provide a foundational understanding, it predates several pivotal developments that have emerged over the last years, such as using RL in the context of large language models (LLMs), e.g., as part of RL from human feedback (RLHF). We further provide a more comprehensive analysis of the reviewed literature across multiple dimensions, namely, used fairness definitions, single- and multi-agent approaches, application domains, and trade-offs, and highlight papers not covered by previous work. Our paper offers an extensive survey of the most recent advancements in fairness in RL and uncovers gaps and potential avenues for future research that have yet to be explored.\\

The rest of the paper is organized as follows. Section \ref{preliminaries} will provide a brief theoretical overview of RL. In Section \ref{def}, we introduce fairness definitions used in RL systems. We then show potential application domains for fair RL systems in Section \ref{appdomains}. Section \ref{methods} follows with a description of the methods used to introduce fairness notions in single- and multi-agent RL systems before we discuss trade-offs in Section \ref{trade}. We conclude our work with an overview of potential future research directions in Section \ref{future}.

\section{Preliminaries} \label{preliminaries}

We start by reviewing the fundamentals of RL. For a more detailed account of RL, we refer the reader to \citet{sutton2018reinforcement}'s work. RL is a process where an agent learns to make decisions by trying different actions to maximize rewards without being explicitly told which actions to take \citep{sutton2018reinforcement}. The most significant aspect of this approach is the balance of exploration and exploitation. RL differs from other methods by starting with a fully interactive agent that has specific goals, can perceive its environment, and take actions. A key challenge is dealing with uncertainty in the environment \citep{sutton2018reinforcement}. Key elements in RL include actions $a$ an agent can take, states $s$ in the environment the agent is interacting in, a policy $\pi$, which guides the agent's behavior, and a reward signal $r$ that the agent receives at some point during its interaction with the environment. The agent aims to maximize the total reward over time. The rewards can be immediate or delayed, depending on the problem formulation (see Sections ~\ref{background:bandits} and ~\ref{background:MDPs}).\\

RL problems can further be formulated as single-objective, where the goal is to maximize a single metric or achieve a specific outcome, or multi-objective setting, where the agent has to balance multiple competing objectives, requiring a more nuanced decision-making process to optimize across various goals. The latter formulation adds a layer of complexity, as the agent must navigate and prioritize these objectives in its decision-making process. Another dimension to categorize RL methods is whether they are using an explicit model of the environment and interaction dynamics. Methods can be model-based, allowing for planning and prediction, or model-free, which relies on trial-and-error learning \citep{sutton2018reinforcement}. Finally, problems can be formulated in RL based on the number of states: bandit problems are RL problems with only a single state, while Markov decision processes (MDPs) are a more general problem formulation that allows for multiple states.

\subsection{Multi-Armed Bandits} \label{background:bandits}

More straightforward RL problems are situations with only one state. These problems do not require learning different actions for different scenarios \citep{sutton2018reinforcement} and can be formulated as a version of the multi-armed bandit (MAB) problem: The agent is repeatedly choosing from $k$ bandits, each yielding a numerical reward based on a fixed probability distribution. The goal is to maximize the total reward over a set number of choices or time steps \citep{sutton2018reinforcement}. This approach, known as nonassociative learning, has been the foundation of most evaluative feedback research. On the contrary, imagine several $k$-armed bandit tasks, each randomly presented but with a unique clue, like a color change in the display, hinting at its identity \citep{sutton2018reinforcement}. This clue does not reveal the (expected) rewards of each bandit but helps to associate each task with its best action; this setting is called an associative or contextual bandit task \citep{sutton2018reinforcement}. It combines trial-and-error learning to find the best actions and the ability to associate these actions with the specific contexts where they are most effective \citep{sutton2018reinforcement}. For an in-depth review of bandit algorithms, we refer the reader to \citet{lattimore2020bandit} and \citet{sutton2018reinforcement}.

\subsection{Markov Decision Processes} \label{background:MDPs}

A more general formulation of RL problems is through MDPs. MDPs are a fundamental framework for sequential decision-making settings, where actions impact not only immediate rewards but also the reachability of future states and, consequently, future rewards \citep{sutton2018reinforcement}. This setting introduces the complexity of balancing immediate and delayed rewards. Unlike bandit problems, where the value of each action is estimated independently, in MDPs, the value is determined for each action within each state. This state-action value, denoted as $Q_{\pi}(s,a)$, is the expected return of an agent, starting from state $s$, taking action $a$, and then following policy $\pi$. Relatedly, the value of a state assuming optimal action selection, $V_{\pi}(s)$, captures the expected value of an agent starting in state $x$ and following policy $\pi$. These notions help attribute long-term strategies' outcomes to specific actions \citep{sutton2018reinforcement}. MDPs frame the challenge of learning through interaction with the environment over multiple time steps to achieve a goal. This interaction is continuous, with the agent choosing actions and the environment responding, presenting new scenarios, and offering rewards \citep{sutton2018reinforcement}. The traditional MDP model can be extended to multi-objective sequential decision-making, where the scalar reward is replaced by a vector whose components represent objectives, also called multi-objective MDPs (MOMDPs) \citep{pirotta2015multi}. The agent aims to maximize these rewards over time through strategic action choices. The distinction between MDPs and bandit problems lies in their complexity and scope. While bandit problems focus on selecting the best action from a set without considering different states or the long-term consequences of actions, MDPs consider the entire sequence of actions and states, focusing on how each decision influences future scenarios and rewards \citep{sutton2018reinforcement}.

\section{Definitions of Fairness in RL} \label{def}
One of the difficulties with studying fairness in RL comes from the numerous definitions of fairness one can adopt. Researchers debated notions of fairness in fields such as philosophy and economics long before fairness in RL – and ML more broadly – were discussed \citep{rawls1971atheory, wolff1998fairness, rabin1993incorporating}, underscoring the complexity and multifaceted nature of the concept. The notions used by ML researchers can be categorized based on where they fit into the ML development cycle (e.g., pre-processing, in-processing, or post-processing fairness) or by their intended aim (focusing on fairness with respect to individuals or groups). We recommend further reading for those interested in comprehensively examining fairness in ML \citep{mehrabi2021survey}. This chapter, however, narrows its focus to the definitions of fairness specifically within the field of RL.\\

\textbf{Welfare-Based Definitions.} \citet{siddique2020learning} and \citet{zimmer2021learning} consider fairness in RL and encode fairness requirements as social welfare functions in the objective function. In both papers, the definition of fairness builds on three components: impartiality, equity, and efficiency. First, impartiality means that all agents are identical. Secondly, equity is based on the \textit{Pigou-Dalton principle}, which states that a reward transfer from a better-off agent to a worse-off agent yields a more desirable utility profile, given that all agents' total reward/utility remains the same. Thirdly, efficiency is the idea that between two potential solutions, if one solution is preferred by all agents, weakly or strictly, then that should be chosen over the other \citep{zimmer2021learning}. This requirement is necessary because it prevents a situation where giving no reward to all users is treated the same as giving non-zero rewards to all users, even though both satisfy the impartiality and equity requirements.
These principles are encoded in social welfare functions (SWF), which measure how good a utility vector is concerning social good. Concretely, a SWF is a function $\phi : \mathbb{R}^D \rightarrow \mathbb{R}$, where $D$ is the number of objectives in \citet{siddique2020learning} and the number of users\footnote{\cite{zimmer2021learning} formulated the problem in terms of ``users", not agents, because a user can represent an individual or a group of individuals, leading to a more general representation.} in \citet{zimmer2021learning}.\\


Despite building on the same three components, it is important to realize the differences in the definitions \citet{siddique2020learning} and \citet{zimmer2021learning} proposed. \citet{siddique2020learning} considers learning fair policies in \textit{single-agent} deep reinforcement learning. Because of the single-agent setting, the notion of ``fairness'' is not across multiple users or agents but across multiple objectives or criteria. Informally, fairness in this single-agent context refers to how the solution ``balances'' the different objectives. The critical assumption in the multi-agent setting in \citet{zimmer2021learning}'s work is that all users are identical, implying that with regard to the impartiality requirement, users should be treated at least similarly. Another important difference is that \citet{siddique2020learning} operationalize these fairness components using the generalized \textit{Gini} social welfare function (GGF), where the GGF is defined as
\begin{equation}
    \text{GGF}_w(v) = \sum^D_{i=1}w_iv^{\uparrow}_i
\end{equation}

where $v \in \mathbb{R}^D$ is the utility vector, $w \in \mathbb{R}^D$ is the fixed positive weight vector, and $v^{\uparrow}$ is the vector with the components of vector $v$ sorted in ascending order \citep{zimmer2021learning}. Depending on the choice of $w$, this approach allows for some flexibility with respect to the exact fairness notion adopted. For example, if $w_1=1$ and all other $w_i=0$ where $i\neq1$, then the implementation corresponds to the maxmin egalitarian notion of fairness \citep{rawls1971atheory}.\\

\textbf{Weighted Proportional Fairness Definition.} \citet{kushner2004convergence} originally defined proportional fairness in the scheduling context to balance two conflicting objectives, namely maximizing the total throughput of a network and guaranteeing a minimum level of service to all users. \citet{liu2020balancing} extended this definition to use weighted proportional fairness as their target fairness metric in the context of interactive recommender systems (RS). The authors first define an allocation vector $x_t^i$ representing the allocation proportion of a group $i$ up to time $t$:
\begin{align}
    x_{t}^{i}=\frac{\sum_{k=1}^{t} y_{a_{k}} \mathbb{I}{\mathcal{A}{c_{i}}}\left(a_{k}\right)}{\sum_{i=1}^{l} \sum_{k=1}^{t} y_{a_{k}} \mathbb{I}{\mathcal{A}{c_{i}}}\left(a_{k}\right)}
\end{align}

where $A$ denotes a group of items with an attribute value $c$ and $i$ refers to the $i$th group. $\mathbb{I}_A(x)$ is 1 if $x \in A$ and 0 otherwise. They further define $y_{a_k}$ as a user's feedback on a recommended item $a_k$. The authors then define weighted proportional fairness as a generalized Nash solution across multiple groups. The weighted proportionally fair allocation is, hence, the solution to the following optimization problem:
\begin{align}
    \max{x_t} \sum_{i=1}^{l} w_{i} \log \left(x_{t}^{i}\right) 
\end{align}
\begin{align*}
    \text { s.t. } \sum_{i=1}^{l} x_{t}^{i}=1, x_{t}^{i} \geq 0, i=1, \ldots, l
\end{align*}
where $w_i$ is a parameter used to weigh the importance of each group. Introducing a weighted component allows for a more differentiated, nuanced notion of fairness that can be adapted to a specific context where there is a need for a nuanced consideration of protected groups.\\



\textbf{Coefficient of Variation.} In multi-agent systems where we want to ensure a fair distribution of resources or rewards among agents, the coefficient of variation ($cv$) is often used to measure fairness among agents \citep{jiang2019learning, elmalaki2021fair}. The metric is defined as follows, capturing the sum of individual differences from the mean: 
\begin{equation}
cv = \sqrt{\frac{1}{n-1} \sum_{i=1}^{n} \frac{\left(u_{i}-\bar{u}\right)^{2}}{\bar{u}^{2}}}
\end{equation}
where $u_i$ is the utility of agent $i$, $n$ is the number of agents, and $\bar{u}$ is the average utility of all agents. The lower $cv$, the fairer the system. \citet{jiang2019learning} note that in multi-agent sequential decision-making, optimizing the coefficient of variation of individual agents is often hard because its value depends on the joint policy of all agents. In these cases, each agent $i$'s contribution to $cv$ is usually approximated by $(u^{i}-\bar{u})^2/\bar{u}^{2}$. The coefficient of variation is then often built into the reward function of each agent so that each agent is punished for getting too much or too little resources or utility \citep{jiang2019learning, elmalaki2021fair}.\\


\textbf{Q-Value Based Definitions.} A natural idea of fairness in RL is based on Q-values. In the MDP setting, the Q-function $Q(s,a)$ is the expected utility of taking a given action at a given state. In this context, $Q^*(s,a)$ is the expected utility of taking action $a$ from state $s$ and (thereafter) acting optimally. Based on this notion, \citet{jabbari2017fairness} introduce the idea of exact fairness and its two relaxations: approximate-choice fairness and approximate-action fairness. Exact fairness requires that in any state $s$, the algorithm never chooses an action $a$ with a higher probability than another action $a'$ unless $Q^*(s, a) > Q^*(s,a')$, i.e., in cases where the long term (discounted) reward of choosing $a$ is higher than that of choosing $a'$. Approximate-choice fairness requires the algorithm to never choose a worse action with a substantially higher probability than better actions. On the other hand, approximate-action fairness requires that an algorithm never favors an action of substantially lower quality than that of a better action. Both exact fairness and approximate-choice fairness require exponential time learning algorithms to approach optimality. Further relaxation to the probabilistic requirement results in a weaker definition of fairness but ensures a polynomial-time learning algorithm.\\





\textbf{$\alpha$-Fair Utility.} 
In computer networking, fairness is often defined as fairly allocating network resources (i.e., bandwidth) to different data flows. In this context, \citet{chen2021bringing} use an $\alpha$-fair utility function to capture fairness notions. For $\alpha \geq 0$, the $\alpha$-fair utility is defined as:
\begin{equation}
U(x)=\left\{\begin{array}{ll}
x^{1-\alpha} /(1-\alpha) & \text { for } \alpha \neq 1 \\
\log (x) & \text { for } \alpha=1
\end{array}\right.
\end{equation}
Note that this notion of fairness is different from the previous notions of fairness in that this measure can be ``toggled'' or controlled to achieve different levels of fairness: Setting $\alpha=0$, for example, leads to throughput maximization, $\alpha=1$ to proportional fairness, and $\alpha\rightarrow\infty$ to max-min fairness \citep{chen2021bringing}. The developer can set the desired fairness parameter $\alpha$. In contrast, definitions like the coefficient of variation are fairness measures embedded into the RL problem itself via the reward function. In these cases, the reward function can account for fairness, even though the programmer cannot precisely control how fair the outcomes are. \citet{hao2023computing} also use $\alpha$-fair utility in the context of computing offloading in edge computing similarly.\\

\textbf{Issue-based Fairness.} \citet{zhu2023fairness} studies fairness in the context of robot assistant services. Compared to the previously discussed notions of fairness, the authors identify issues in human-robot interactions and define fairness as different possibilities that these issues present given groups of people with and without sensitive identities. They identified 
\begin{itemize}
    \item \textit{willingness issues}, i.e., the issue that a robot is more likely to respond to groups without sensitive identities
    \item \textit{quality issues}, i.e., the issue that a robot is more likely to be at an uncomfortable distance from groups with sensitive identities
    \item \textit{priority issues}, i.e., the issue that a robot is more likely to prioritize groups without sensitive identities
    \item \textit{risk issues}, i.e., the issue that a robot is more likely to be at a risky distance from groups with sensitive identities 
\end{itemize} 

The authors calculate the issue-specific score as $d\left(g_c\right)=\left|\operatorname{Pr}\left(R=r \mid P \in g_c\right)-\operatorname{Pr}\left(R=r \mid P \notin g_c\right)\right|$. They then calculate an \textit{overall issue score} as the average of the scores of all issue-specific scores; this average issue score is subsequently included in their approach as a proxy for fairness (see Section ~\ref{methods} for more details).\\

\textbf{Calibrated Fairness.} \citet{liu2017calibrated} introduce smooth and calibrated fairness as two notions of fairness in sequential decision-making problems. Smooth fairness ensures that if two groups (or arms in the context of multi-armed bandit problems) have similar quality distributions, they should be selected with similar probabilities, addressing the issue of minor quality differences unfairly influencing selection. Calibrated fairness, on the other hand, demands that the probability of selecting a group matches the probability that this group has the best quality realization. This concept ensures meritocracy by requiring that dissimilar groups be treated in accordance with their differences in quality, thus addressing both the need for equality among similarly qualified groups and the requirement to recognize and reward merit when differences are significant.\\

\textbf{Regularized Maximin Fairness.} \citet{zhang2014maximinfairness} introduce a new fairness criterion specifically designed for multi-agent MDPs, where agents may have conflicting interests. The core idea behind this fairness criterion is to enhance equity among agents by focusing on the least advantaged positions. It seeks to maximize the minimum (or worst) performance across all agents, thereby ensuring that the most disadvantaged agent's outcome is improved as much as possible. This maximization is not done in isolation but is regularized with an overall performance consideration, aiming for a balanced approach that uplifts the worst-off agent without excessively compromising the collective outcome. This approach is reflective of a Pareto-efficient solution, where improvements can be made to at least one agent's situation without making any other agent worse off \citep{zhang2014maximinfairness}.\\

\textbf{Observations on Fairness Definitions.}
In general, we notice that definitions of fairness in RL are highly inconsistent, some of them are even mutually exclusive, which has been previously discussed in the literature \citep{berk2017convex, friedler2016possibility}. This inconsistency arises due to several reasons. First, the majority of the existing literature is focused on incorporating fairness in RL in one specific application domain. For example, the work of \citet{elmalaki2021fair} focuses on incorporating fairness in RL for IoT devices, while \citet{chen2021bringing} studies fairness in RL for network utility optimization. This approach results in many fairness approaches being highly specialized to certain contexts that may not work well in more general, context-agnostic RL settings. Second, many notions of fairness originated from fields outside of ML. For example, the notion of $\alpha$-fairness was originally studied in the context of networks while welfare economics definitions have their origin in social welfare and game theory. This causes a discrepancy in definitions adopted in RL, since there has neither been an agreement on a general fairness definition in other fields nor in ML or RL – and there likely never will be, given that the most suitable fairness definition strongly depends on compromises and context of application. However, a structured approach to decide on a fairness definition to be used in a specific RL application is not available, either, leading to often arbitrary choices of fairness definitions, without sufficient justification. While \citet{mandal2022socially} outline four axioms that a fairness measure should satisfy, there's more work necessary on how to decide which fairness measure is appropriate given an RL problem.

\section{Application Domains} \label{appdomains}
Fairness in RL is most relevant in cases where multiple stakeholders use a system, or its decisions will impact multiple users. Ensuring fairness, or at least the attempt to ensure fairness under generally accepted definitions, is crucial for the users' fair treatment and acceptance of these systems. This chapter presents specific domains in which fair RL has been studied.\\

\textbf{Recommendation Systems.} \citet{liu2020balancing} developed a model to balance fairness and accuracy in interactive RS. These systems are being used to recommend items of interest (e.g., news, movies, or articles \citep{steck2015interactive}) to individual users. The recommendations are updated in an online setting based on user feedback, often expressed as taking a desired action or not, such as buying an article after seeing a recommendation. The latter is also known as conversion rate. Only using the conversion rate as the single objective can lead to an imbalance across different demographic groups, which can lead to minorities being ignored in recommendations. \citet{ge2022toward} identified the same issue in RS and used a fairness-aware MOMDP approach to optimally balance these two objectives (see Sec.~\ref{methods}). Other work by \citet{huang2022contextual} focused on fair personalized RS utilizing contextual bandits while \citet{huang2022causal} used causal bandits for fair recommendations. Similar work on fair RL for recommendations was done by \citet{singh2020healthyrecommendation} and \citet{wang2021fairnessexposure}. For a broader overview on fairness in RS, we refer the reader to \citet{li2023fairness} and \citet{zhang2021recommendationfairness}.\\ 

\textbf{Work Distribution.} \citet{claure2019reinforcement} looked at fair RL in the context of resource distribution in human-robot teams. They specifically focused on a fair candidate selection for a work task because in an unconstrained RL setting, the agent would learn the individual worker's performance first and then assign as many tasks as possible to the highest-performing candidate. In real life, this can lead to a two-fold issue: Firstly, when a human worker is allocated significantly more resources to process than his team members, he might burn out quickly, which is a factor that is not being taken into account by the agent by default. Secondly, favoring one employee might hurt all team members' motivation. Perceived inequalities have been shown to motivate people to act against their self-interest to eliminate the inequality \citep{camerer2003behavioral}, sometimes including actions to retaliate \citep{skarlicki1997retaliation}. It further undermines their trust in the system due to a perceived unfairness. All these aspects can negatively impact overall performance if fairness is not explicitly considered in the system.\\


\textbf{Scheduling and Resource Distribution.} \citet{chen2021bringing} look at fair RL solutions in the context of wireless network scheduling and Quality-of-Experience (QoE) optimization in video streaming. The former is concerned with scheduling the transmission and reception of sequences of network packages across a network of users. In this case, fairness considerations are necessary to prevent the discrimination of certain parties and, in the worst case, the restriction of access to the network. \citet{yamazaki2021congestion} also look into fair RL for wireless network scheduling. Fairness regarding QoE, on the other hand, means that all users have a similar, e.g., video streaming experience based on metrics to capture the (perceived) video quality by users. Fair RL in the context of QoE has also been addressed by \citet{arani2021fairness, tong2021qoefairness} and \citet{petrangeli2014multi}. Similarly, \citet{hao2023computing} addressed the challenge of computing offloading in edge computing, where balancing service provision at network edges with limited resources is crucial for user experience. Other work that pertains to fair RL in scheduling and resource distribution was done by \citet{agarwal2019reinforcement}, \citet{comsa2019comparison}, and \citet{arianpoo2016network}. Another area in this context is Unmanned Aerial Vehicles (UAV) Control. UAVs serve as dynamic flying communication platforms, enabling quick and flexible deployment with the ability to establish direct Line-of-Sight links, yet face significant challenges such as limited battery life and coverage area, alongside high costs \citet{qi2020uavcontrol}. RL can be employed to optimize UAV control for efficient coverage and communication, ensuring fairness in resource allocation across the targeted regions despite these constraints \citep{qi2020uavcontrol}. This setting has further been explored by \citet{liu2018energy}.\\

\textbf{Infrastructure Control.} \citet{siddique2020learning} and \citet{zimmer2021learning} consider applications to traffic light control and data center control. In the data center control setting, the goal is to minimize the queue length of each network switch, which must be fair. In the traffic light control problem, without fairness constraints, the problem seeks to minimize the total wait time across all lanes. By incorporating fairness considerations, the agent is then tasked to learn how to optimize the expected wait time per road. \citet{athalye2021fairness, raeis2021trafficcontrol, valkanis2022traffic} and \citet{li2020fairness} also looked into fair RL-based ground traffic control. \citet{de2010fairness} use multi-agent RL for fairness-aware air traffic flow control.\\

\textbf{Internet-of-Things (IoT).} In the area of IoT, fairness considerations could enable IoT devices to mitigate better the challenges brought by intra-human, inter-human, and multi-human variability. Take smart thermometers, for example. The same user's temperature preference might change over time (intra-human variability), each user has different temperature preferences (inter-human variability), and multiple people in the same room can have a wide range of temperature preferences (multi-human variability). \citet{elmalaki2021fair} show that using their developed RL-based ``FaiR-IoT" fairness-aware human-in-the-loop framework will improve the user experience and improve fairness (measured by the coefficient of variation, discussed in Sec~\ref{def}).\\

\textbf{Robotics.} A key area for the use of RL is to train robots to interact in the real world. For example, robots can be used to provide services in airports or restaurants to humans; fair RL in this context was studied by \citet{zhu2023fairness} to ensure that humans impacted by the robot are treated fairly. On the other hand, \citet{zhu2018multirobot} researched fairness pertaining to RL-based multi-robot coordination.\\

\textbf{Other Areas.} For completeness, we also want to mention less prominent application areas for fair RL here. For example, \citet{atwood2019fair}  studied the precision disease control problem, i.e., the optimal allocation of vaccines in a social network, using fair RL. \citet{bao2019fairness} used a multi-agent RL approach for fair stock trading. Finally, \citet{wang2020facialrecognition} used RL to mitigate bias in facial recognition.


\section{Methodology} \label{methods}
Current literature on fairness in RL can be categorized into research that focuses on single- or multi-agent setups. We will discuss these approaches in the following chapters. For a general overview on multi-agent RL, we refer the reader to \citet{busoniu2006multi}.



\subsection{Single-agent RL}




In the single-agent multi-objective setting studied by \citet{siddique2020learning}, the authors integrate the GGF (see Sec.~\ref{def}) into their MOMDP formulation. The problem of solving for an optimal policy that ensures a fair distribution of rewards to $D$ users then becomes 
\begin{equation}
\text{argmax}_{\pi}\text{GGF}_w(J(\pi))
\end{equation}
where $J(\pi)$ is the discounted reward \citep{siddique2020learning}. Note that because the GGF is a non-linear function, this is a non-linear convex optimization problem \citep{siddique2020learning}. To solve it, a modified Deep Q Network (DQN) or policy gradient methods could be used. The latter can be advantageous because they directly optimize for the desired objective function and can also learn stochastic policies instead of just deterministic ones.\\

\citet{liu2020balancing} propose the RL-based \textit{FairRec} framework to ensure a balanced long-term trade-off between accuracy and fairness in a single-agent MDP setup in RS. The authors use the weighted proportional fairness notion described in Sec.~\ref{def}. \cite{liu2020balancing} build an actor-critic architecture where the actor network is responsible for dynamic recommendations depending on the fairness status and the user preferences. On the other hand, the critic network encourages or discourages a recommended item based on its value estimate of the actor's output. To accomplish a focus on both accuracy and fairness, \citet{liu2020balancing} design a two-fold reward based on a personalized fairness-aware state representation. They consider whether a user performed a desired activity on a recommended item before they evaluate the fairness gain of the activity.\\ 


\citet{ge2022toward} introduce a fairness-aware recommendation framework called MoFIR, which utilizes a MOMDP setting to find an optimal balance between conflicting objectives of utility and fairness. MoFIR is designed to learn a single parametric representation that can adapt to different preferences over these objectives, accommodating the varying needs of real-world e-commerce platforms \citep{ge2022toward}. The key innovation in MoFIR is the modification of the Deep Deterministic Policy Gradient (DDPG) algorithm by incorporating conditioned networks (CN) \citep{ge2022toward}. These networks are conditioned directly on the preferences for utility and fairness, allowing the system to output Q-value vectors that represent the optimal policy for any given preference mix \citep{ge2022toward}. Constrained optimization problem formulations as by, e.g., \citet{nabi2019learning}, aim to maintain a certain level of fairness while optimizing for the main performance objective. However, these approaches often compromise recommendation accuracy for fairness. \citet{ge2022toward} overcome this limitation with their approach.\\

\citet{claure2019reinforcement}, on the other hand, consider a stochastic MAB framework together with an unconstrained upper confidence bound (UCB) algorithm in their work. The aim for an agent in the context of an MAB is to maximize cumulative rewards by pulling bandits' arms based on previous information the agent obtained. In the UCB algorithm, the agent does this by using the information on the number of times an arm was pulled and the average empirical rewards the agent received to estimate the expected reward of each arm. However, the authors find that an unconstrained UCB algorithm fails to ensure fairness since the agent will not use under-performing arms after a certain number of time steps. To prevent this, \citet{claure2019reinforcement} propose two adjusted UCB algorithms: a strict-rate-constrained and a stochastic-rate-constrained version. The former guarantees\footnote{See \citet{claure2019reinforcement} for formalized theoretical guarantees.} that each lever has a minimum pull rate because arms are pre-scheduled to be pulled in fixed time slots. In all other time slots, the agent will follow the standard UCB algorithm, i.e., choose the best arm according to the benchmark strategy. The stochastic-rate-constrained UCB algorithm, on the other hand, only guarantees\footnote{See \citet{claure2019reinforcement} for formalized theoretical guarantees.} that the expected pulling rate is at least the minimum pull rate at any time. Compared with the strict-rate-constrained UCB algorithm, the authors introduce randomness, thereby ensuring that the probability that the agent pulls an arm at time t is equivalent to the minimum pull rate. In this case, with a probability of $1-Kv$, where $K$ is the arm set and $v$ is the minimum pull rate, the benchmark UCB policy is followed.\\

Looking at other fairness in MAB problem formulations, \citet{chen2020fair} utilize contextual information about users and tasks, allowing their algorithm to dynamically adjust its allocation strategy to satisfy fairness constraints while still aiming to maximize a given performance metric. This algorithm uniquely does not assume a fixed distribution for the performance losses associated with each user, as did previous related work by \citet{patil2021achieving, li2019combinatorial} and \citet{claure2019reinforcement}, making it adaptable to non-stationary and adversarial environments. \citet{ghodsi2021online} also explore fairness in MAB settings. The authors employ the EXP\_3 algorithm to address the online cake-cutting problem with non-contiguous pieces. This approach allows for sub-linear fairness and revenue regret by balancing exploration and exploitation in an environment where agents' valuations change. The authors' methodology involves running two separate algorithms independently to achieve sub-linear regret for revenue and fairness, then merging these algorithms to tackle both objectives simultaneously. Finally, \citet{gillen2018online} present an algorithm for online learning in linear contextual bandits settings with individual fairness constraints governed by an unknown Mahalanobis similarity metric. The algorithm learns from weak feedback identifying fairness violations without quantifying their extent, akin to regulator interventions that can recognize unfairness but cannot define it quantitatively. The main result is an adversarial-context algorithm that limits the number of fairness violations logarithmically over time $T$ while obtaining optimal regret bounds concerning the best fair policy.

\subsection{Multi-agent RL}

In the context of multi-agent fair RL, \citet{chen2021bringing} take the idea by \citet{liu2020balancing} to the multi-agent setting and aim to optimize general fairness utility functions in actor-critic RL. They specifically developed a method to adjust the standard RL rewards by a multiplicative weight that considers the history of rewards and the shape of the fairness utility. The multiplicative adjustment is defined using a uniformly continuous function $\Phi(h_{\pi,t})$ that is dependent on a statistic $h_{\pi,t}$ capturing past rewards. The adjusted rewards are calculated using $\hat{r}_{k, t}=r_{k, t} \cdot \phi\left(h_{\pi, t}\right)$.\\

The authors then employ a rewards-adjusted actor-critic architecture to address the issue that a non-linear $\alpha$-fair utility ($\forall \alpha > 0$) does not satisfy the Markovian property, which would be necessary to formulate $\alpha$-fair network utility optimization as an MDP and guarantee convergence under the Policy Gradient Theorem \citep{chen2021bringing}. Instead, their approach guarantees that, given a proper choice of $\Phi$ and $h$, it converges to ``at least a stationary point of the $\alpha$-fair utility optimization'' \citep{chen2021bringing}. Another advantage to this approach is that it builds on actor-critic architectures, meaning that the optimization converges quicker because of variance reduction and because the proposed algorithm does not rely on the Monte Carlo method, especially when optimizing in large state/action spaces.\\

\citet{agarwal2019reinforcement} propose both model-based and model-free algorithms to optimize a non-linear function of long-term average rewards in multi-agent contexts. Their model-based algorithm utilizes posterior sampling with a Dirichlet distribution, showing convergence to an optimal point under certain conditions and achieving a sub-linear regret bound in the number of time steps and objectives. The paper also introduces a model-free policy gradient algorithm that can be implemented efficiently using neural networks. These methods are designed to overcome the limitations of traditional RL approaches, such as SARSA, Q-Learning, and DQN, which are sub-optimal for non-linear, multi-agent optimization problems.\\

\citet{arianpoo2016network} use a distributed, cross-layer approach leveraging Q-learning to monitor network dynamics and adjust TCP parameters autonomously for fair resource allocation. Each node hosting a TCP source models the system's state as an MDP. This approach allows each node to autonomously adjust TCP parameters without central control or the need for control message exchanges among nodes. The authors emphasize the suitability of Q-learning for this application due to its model-free nature, enabling it to find optimal strategy-selection policies for any given finite state MDP, particularly under the dynamic conditions of wireless networks.\\


In the multi-agent setup researched by \citet{zimmer2021learning}, the authors formulate the problem of cooperative multi-agent RL as 
\begin{equation}
    \max_{\theta} \phi(J(\theta))
\end{equation}
where $\theta$ is the joint policy of all the agents and $J_k(\theta) = E_{\theta}[\sum_k \gamma^t r_{k, t}]$ the expected sum of discounted rewards of user $k$, and $J(\theta) = \sum_k J_k(\theta)$. The algorithmic solution for this optimization problem includes a policy gradient approach implemented in an actor-critic architecture. \citet{zimmer2021learning} propose Self-Oriented Team-Oriented (SOTO) networks updated by a dedicated policy gradient. On a high level, the network includes a self-oriented and a team-oriented policy. The former optimizes for an individual policy, whereas the team-oriented policy optimizes for the SWF $\phi(J(\theta))$, i.e., the two sub-networks focus on efficiency and equity, respectively. An advantage of the approach by \citet{zimmer2021learning} is that it is not domain-specific and allows for the adoption of a variety of fairness notions since it only demands a (sub-)differentiable welfare function \citep{zimmer2021learning}.\\

\citet{jiang2019learning}, on the other hand, introduce the Fair-Efficient Network (FEN), an RL-based model that makes use of a ``fair-efficient reward'' \citep{jiang2019learning} to address multi-agent RL settings. Each agent learns this reward to optimize its own policy. Additionally, an average consensus among agents as part of the fair-efficient rewards allows coordination between agents' policies. Specifically, the authors consider a setting with $n$ agents and limited, commonly-accessible resources. Each agent's fair-efficient reward at time t is:

\begin{equation}
\hat{r}_{t}^{i}=\frac{\bar{u}_{t} / c}{\epsilon+\left|u_{t}^{i} / \bar{u}_{t}-1\right|}
\end{equation}

where the utility of agent $i$ at time step $t$ is the average reward over time it received, i.e., $
u_{t}^{i}=\frac{1}{t} \sum_{j=0}^{t} r_{j}^{i}$, and $c$ is the maximum environmental reward an agent can obtain at a time step. Hence, $\bar{u}_{t} / c$ can be seen as a proxy for system efficiency since it represents the system-wide resource allocation. On the other hand, $\left|u_{t}^{i} / \bar{u}_{t}-1\right|$ represents an agent's utility deviation from the average, which is taken as a proxy for fairness based on the idea of the coefficient of variation (see Sec.~\ref{def}). Each agent uses this reward to learn its own policy $F_{i}=\mathbb{E}\left[\sum_{t=0}^{\infty} \gamma^{t} \hat{r}_{t}^{i}\right]$ with a discount $\gamma$. Potential multi-objective conflicts are circumvented by using a hierarchical RL model with a controller that maximizes the fair-efficient reward by changing between multiple sub-policies. These sub-policies are designed with respect to different goals: to maximize an environmental reward and to explore different fair behaviors. This approach aims to enable agents to learn efficiency and fairness simultaneously.\\

Similarly, \citet{zhu2023fairness} study a multi-agent setting and employ a self-reflecting approach that allows agents to ``self-identify biased behavior during interactions with humans'' \citep{zhu2023fairness}; they call this Fairness-Sensitive Policy Gradient Reinforcement Learning (FSPGRL). Based on the issue score $I$ explained in Sec.~\ref{def}, the authors define a reward penalty parameter $\tau_{penalty}$ that captures the tolerance for the identified biases. They then define the reward as $R_t=\tau_{\text{penalty}} \sum\left(1-I\left(g_c\right)\right)$, where $g_c$ is the group of people with sensitive attributes. They combine the REINFORCE algorithm, a policy gradient method, and Proximal Policy Optimization (PPO) to mitigate the bias and allow the agent to self-correct biased behavior. The agent itself is based on an actor-critic network. FSPGRL subsequently ``identifies bias by examining the abnormal update along particular gradients and updates the policy network to support fair decision-making'' \citep{zhu2023fairness}.\\

\citet{grupen2022cooperative} introduce a different multi-agent angle to fair RL: team-based fairness, a group-based fairness measure that seeks equitable reward distributions across sensitive groups within a team. The authors subsequently introduce Fairness through Equivariance, a strategy that transforms the team's joint policy into an equivariant map to enforce team fairness during policy optimization. They further introduce Fairness through Equivariance Regularization, a soft-constraint version of Fairness through Equivariance, which allows for tuning fairness constraints over multi-agent policies through regularization. The latter aims to balance fairness and utility by adjusting the weight of equivariance regularization.\\

Finally, \citet{hao2023computing} formulate the setting of computing offloading in edge computing as a long-term average optimization problem, aiming to maximize an $\alpha$-fair utility function under computing capacity and storage constraints and approach it as an MDP. To solve it, they propose a multi-update deep RL algorithm that optimizes for fairness and efficiency.



\section{Trade-Offs} \label{trade}
Imposing fairness requirements to RL algorithms often results in worse running time compared to problems without fairness constraints. \citet{jabbari2017fairness} show the trade-offs between efficiency and fairness. In particular, both fair and approximate-choice fair requirements impose an exponential time step $\mathcal{T} =\Omega(k^n)$. The authors' defined approximate-action fairness (see Sec.~\ref{methods}) requires steps $\mathcal{T} =\Omega(k^{1/(1-\gamma)})$, where $n$ is the size of the state space and $\gamma$ is the discount factor. Note that without fairness constraints, standard RL algorithms learn an $\epsilon$-optimal policy in a number of steps polynomial in $n$, $1/\epsilon$, and all parameters of the MDP. This example shows that imposing fairness requirements potentially comes with a cost regarding run-time efficiency.\\

Imposing fairness requirements further decreases performance. \citet{liu2020balancing} discussed this trade-off in their paper. The authors find that their \textit{FairRec} framework (see Sec.~\ref{methods}) does increase fairness in RS while maintaining a good recommendation quality but that, compared to non-fair baselines, there are still performance losses. \citet{ge2022toward} also identified this trade-off as an issue; by proposing a Pareto-optimal approach, they aim to provide optimal fairness-performance combinations and give users the option of choosing whichever optimal performance-fairness combination suits their application context best. The trade-off between fairness and performance is a general issue in designing fair RL systems since any fair solution will introduce additional constraints and/or objectives that will, compared to an unconstrained, non-fair problem formulation, result in a decreased performance quality \citep{berk2017convex}. This trade-off is an ongoing challenge in ML more broadly \citep{wang2021tradeoff}. However, since the perceived fairness of a system can impact human performance or trust, not considering fairness considerations can lead to an overall sub-optimal performance \citep{elmalaki2021fair}. To better understand these trade-offs, \citet{cimpean2024group} develop a performance-fairness trade-off framework that allows users to explore ``obtainable performance-fairness trade-offs, [that] can be used by stakeholders to select the best policy for the problem at hand.'' \citep{cimpean2024group},


\section{Future Research Directions} \label{future}

The study of fairness in RL has been and will continue to be an interdisciplinary field involving economics, mathematics, computer science, philosophy, and political science. As the deployment of RL-enabled systems accelerates,  the different communities need to address the following gaps and challenges. \\

\textbf{Fairness in RLHF.} RLHF is a method for optimizing language models based on human feedback by integrating and preferences into the learning process \citep{kaufmann2023survey}. Traditionally, this is being done by having a single reward model that is derived from the given preference data \citep{chakraborty2024maxmin}. In implementing fairness in the context of RLHF, develoeprs must consider three dimensions: 1) the representation of the human feedback providers to encompass a broad range of demographics, cultures, and ethical viewpoints, 2) how these differing preferences can fairly be integrated to fine-tune the reward mechanisms of the agent, and 3) ensuring that the subsequent model not only maximizes its primary objectives but also takes into account a fairness objective. All three dimensions are understudied. On a conceptual level, \citet{liu2023transforming} looked at how RLHF can influence society more broadly. Beyond that, \citet{zhong2024provable} published theoretical work on explicitly modelling diverse perspectives and integrating fairness concerns in an offline RL setting. Future work is necessary to understand the empirical performance and scalability of \citet{zhong2024provable}'s work along with research into incorporating fairness in a multi-party RLHF in an online setting, as well as extending the analysis to more general function approximation classes for modeling individual rewards. Finally, \citet{chakraborty2024maxmin} propose MaxMin-RLHF, which learns a mixture of reward models to capture diverse human preferences and optimizes a max-min objective over user group utilities to achieve socially fair alignment of language models. Important extensions of this work would be evaluating \citet{chakraborty2024maxmin}'s approach on real-world human preference data with organic diversity instead of just synthetic preference data, as well as exploring alternative social welfare objectives.



\textbf{Cross-Domain Fair RL Approach.} As discussed in Sec.~\ref{def}, the diversity of fairness definitions in the field of fair RL are considerable. In most works we surveyed, the definitions were derived from domain-specific requirements. To advance fair RL algorithms, we propose investigating the merits and drawbacks of each of these definitions to find a more general definition that can be applied across domains or a structured approach to determining which fairness definition is appropriate for a specific context. To do so, one could, for example, expand the approach by \cite{zimmer2021learning}, which accepts different (sub-)differentiable notions of fairness, and test the effects of different fairness definitions on the performance of subsequent models. Defining fairness across contexts is a more broad open-problem in ML, so making progress in either ML or RL could advance the field. Frameworks that can be used with multiple fairness definitions could also help to study the trade-offs between fairness definitions or achieve multiple fairness requirements at the same time\footnote{This cannot be achieved for all known fairness definitions since some of them are mutually exclusive as shown by \citet{miconi2017impossibility}.}; e.g., \citet{cimpeantowards} outline a fairness-definition-agnostic framework but do not validate it or show its effectiveness. In the bandit context, \citet{metevier2019offline}'s work is promising for accepting multiple fairness definitions in an offline contextual bandit algorithm. A related open challenge is fair RL methods that can be applied across domains. While developing domain-specific fair RL algorithms ensures a high degree of applicability to the respective context, we recommend investigating more general models that can be applied across domains. The approaches taken by \cite{zimmer2021learning}, \cite{liu2020balancing}, and \cite{chen2021bringing} are promising in this direction but should be tested for effectiveness in a more diverse set of domains to verify their applicability and robustness. Finally, another interesting approach that has not extensively been explored in an RL context yet is personalized fairness \citep{li2021towards}, where users can articulate their unique expectations of fairness. This approach prevents the system from solely relying on the designers' interpretations to navigate through conflicting definitions of fairness.\\

\textbf{Additional Application Domains.} While we advocate for the focus on general cross-domain fair RL algorithms, we further encourage research in application-specific settings to find optimal, context-specific models, especially in high-stakes contexts to ensure the highest possible performance. Potential domains to investigate are routing, traffic light control systems, and cloud computing \citep{jiang2019learning}.\\

\textbf{Fairness over Time.} In the existing fair-RL literature, fairness has often been treated as a final objective to be fulfilled or maximized at the final time step $T$. However, ensuring fairness at any point during the interaction should become a research focus, especially for continuous RL applications. For example, the research by \citet{henzinger2023runtime} could be expanded to the RL setting. We further recommend studying the differences between ensuring fairness at a given time and ensuring fairness in long-term decision-making settings. Specifically, approaches to guarantee fairness not only at the final time step $T$ but throughout the runtime should be investigated.\\

\textbf{Adversarial Fairness.} As RL systems are increasingly used in potentially adversarial settings, the imperative to design learning policies that are both fair and robust becomes critical. Hence, future research should focus on developing approaches that can identify and counteract bias-inducing adversarial attacks while also ensuring equitable treatment of all entities within an RL setting.


\section{Conclusion} \label{concl}

Given the increased adoption of RL in socio-technical systems, supporting – or even guaranteeing – fairness in these algorithms is more important than ever. In this paper, we discussed research that focuses on developing fair RL approaches. We compared different definitions of fairness, showcased the methodologies pursued in the different papers, and proposed future research directions. With our survey, we hope to build a discussion base and encourage further research in the field.

\newpage

\appendix

\bibliography{aaai24}

\begin{thebibliography}{77}
\providecommand{\natexlab}[1]{#1}

\bibitem[{Agarwal and Aggarwal(2019)}]{agarwal2019reinforcement}
Agarwal, M.; and Aggarwal, V. 2019.
\newblock Reinforcement learning for joint optimization of multiple rewards.
\newblock \emph{arXiv preprint arXiv:1909.02940}.

\bibitem[{Arani, Hu, and Zhu(2021)}]{arani2021fairness}
Arani, A.~H.; Hu, P.; and Zhu, Y. 2021.
\newblock Fairness-aware link optimization for space-terrestrial integrated networks: A reinforcement learning framework.
\newblock \emph{IEEE Access}, 9: 77624--77636.

\bibitem[{Arianpoo and Leung(2016)}]{arianpoo2016network}
Arianpoo, N.; and Leung, V.~C. 2016.
\newblock How network monitoring and reinforcement learning can improve tcp fairness in wireless multi-hop networks.
\newblock \emph{EURASIP Journal on Wireless Communications and Networking}, 2016: 1--15.

\bibitem[{Athalye and Nayak(2021)}]{athalye2021fairness}
Athalye, A.; and Nayak, S. 2021.
\newblock Fairness and robustness of mixed autonomous traffic control with reinforcement learning.

\bibitem[{Atwood et~al.(2019)Atwood, Srinivasan, Halpern, and Sculley}]{atwood2019fair}
Atwood, J.; Srinivasan, H.; Halpern, Y.; and Sculley, D. 2019.
\newblock Fair treatment allocations in social networks.
\newblock \emph{arXiv preprint arXiv:1911.05489}.

\bibitem[{Bao(2019)}]{bao2019fairness}
Bao, W. 2019.
\newblock Fairness in multi-agent reinforcement learning for stock trading.
\newblock \emph{arXiv preprint arXiv:2001.00918}.

\bibitem[{Berk et~al.(2017)Berk, Heidari, Jabbari, Joseph, Kearns, Morgenstern, Neel, and Roth}]{berk2017convex}
Berk, R.; Heidari, H.; Jabbari, S.; Joseph, M.; Kearns, M.; Morgenstern, J.; Neel, S.; and Roth, A. 2017.
\newblock A convex framework for fair regression.
\newblock \emph{arXiv preprint arXiv:1706.02409}.

\bibitem[{Bommasani et~al.(2021)Bommasani, Hudson, Adeli, Altman, Arora, von Arx, Bernstein, Bohg, Bosselut, Brunskill et~al.}]{bommasani2021opportunities}
Bommasani, R.; Hudson, D.~A.; Adeli, E.; Altman, R.; Arora, S.; von Arx, S.; Bernstein, M.~S.; Bohg, J.; Bosselut, A.; Brunskill, E.; et~al. 2021.
\newblock On the opportunities and risks of foundation models.
\newblock \emph{arXiv preprint arXiv:2108.07258}.

\bibitem[{Busoniu, Babuska, and De~Schutter(2006)}]{busoniu2006multi}
Busoniu, L.; Babuska, R.; and De~Schutter, B. 2006.
\newblock Multi-agent reinforcement learning: A survey.
\newblock In \emph{2006 9th International Conference on Control, Automation, Robotics and Vision}, 1--6. IEEE.

\bibitem[{Camerer(2003)}]{camerer2003behavioral}
Camerer, C.~F. 2003.
\newblock Behavioral game theory: Plausible formal models that predict accurately.
\newblock \emph{Behavioral and Brain Sciences}, 26(2): 157--158.

\bibitem[{Chakraborty et~al.(2024)Chakraborty, Qiu, Yuan, Koppel, Huang, Manocha, Bedi, and Wang}]{chakraborty2024maxmin}
Chakraborty, S.; Qiu, J.; Yuan, H.; Koppel, A.; Huang, F.; Manocha, D.; Bedi, A.~S.; and Wang, M. 2024.
\newblock MaxMin-RLHF: Towards Equitable Alignment of Large Language Models with Diverse Human Preferences.
\newblock \emph{arXiv preprint arXiv:2402.08925}.

\bibitem[{Chen, Wang, and Lan(2021)}]{chen2021bringing}
Chen, J.; Wang, Y.; and Lan, T. 2021.
\newblock Bringing Fairness to Actor-Critic Reinforcement Learning for Network Utility Optimization.
\newblock In \emph{IEEE INFOCOM 2021-IEEE Conference on Computer Communications}, 1--10. IEEE.

\bibitem[{Chen et~al.(2020)Chen, Cuellar, Luo, Modi, Nemlekar, and Nikolaidis}]{chen2020fair}
Chen, Y.; Cuellar, A.; Luo, H.; Modi, J.; Nemlekar, H.; and Nikolaidis, S. 2020.
\newblock The fair contextual multi-armed bandit.
\newblock In \emph{Proceedings of the 19th International Conference on Autonomous Agents and MultiAgent Systems}.

\bibitem[{Cimpean et~al.(2024)Cimpean, Jonker, Libin, and Nowe}]{cimpean2024group}
Cimpean, A.; Jonker, C.~M.; Libin, P. J.~K.; and Nowe, A. 2024.
\newblock A Group And Individual Aware Framework For Fair Reinforcement Learning.
\newblock In \emph{The Sixteenth Workshop on Adaptive and Learning Agents}.

\bibitem[{Cimpean et~al.(2023)Cimpean, Libin, Coppens, Jonker, and Now{\'e}}]{cimpeantowards}
Cimpean, A.; Libin, P.; Coppens, Y.; Jonker, C.; and Now{\'e}, A. 2023.
\newblock Towards Fairness In Reinforcement Learning.
\newblock In \emph{Proceedings of the Adaptive and Learning Agents Workshop (ALA 2023)}, 1--5.

\bibitem[{Claure et~al.(2019)Claure, Chen, Modi, Jung, and Nikolaidis}]{claure2019reinforcement}
Claure, H.; Chen, Y.; Modi, J.; Jung, M.; and Nikolaidis, S. 2019.
\newblock Reinforcement learning with fairness constraints for resource distribution in human-robot teams.
\newblock \emph{arXiv preprint arXiv:1907.00313}.

\bibitem[{Comșa et~al.(2019)Comșa, Zhang, Aydin, Kuonen, Trestian, and Ghinea}]{comsa2019comparison}
Comșa, I.-S.; Zhang, S.; Aydin, M.; Kuonen, P.; Trestian, R.; and Ghinea, G. 2019.
\newblock {A comparison of reinforcement learning algorithms in fairness-oriented OFDMA schedulers}.
\newblock \emph{Information}, 10(10): 315.

\bibitem[{Corbett-Davies and Goel(2018)}]{corbett2018measure}
Corbett-Davies, S.; and Goel, S. 2018.
\newblock The measure and mismeasure of fairness: A critical review of fair machine learning.
\newblock \emph{arXiv preprint arXiv:1808.00023}.

\bibitem[{de~Arruda et~al.(2010)de~Arruda, Leite, de~Almeida, Crespo, and Weigang}]{de2010fairness}
de~Arruda, A.~C.; Leite, A.~F.; de~Almeida, C.~R.; Crespo, A.~M.; and Weigang, L. 2010.
\newblock Fairness analysis with cost impact for Brasilia's Flight Information Region using reinforcement learning approach.
\newblock In \emph{13th International IEEE Conference on Intelligent Transportation Systems}, 539--544. IEEE.

\bibitem[{Elmalaki(2021)}]{elmalaki2021fair}
Elmalaki, S. 2021.
\newblock FaiR-IoT: Fairness-aware Human-in-the-Loop Reinforcement Learning for Harnessing Human Variability in Personalized IoT.
\newblock In \emph{Proceedings of the International Conference on Internet-of-Things Design and Implementation}, 119--132.

\bibitem[{Friedler, Scheidegger, and Venkatasubramanian(2016)}]{friedler2016possibility}
Friedler, S.~A.; Scheidegger, C.; and Venkatasubramanian, S. 2016.
\newblock On the (im) possibility of fairness.
\newblock \emph{arXiv preprint arXiv:1609.07236}.

\bibitem[{Gajane et~al.(2022)Gajane, Saxena, Tavakol, Fletcher, and Pechenizkiy}]{gajane2022survey}
Gajane, P.; Saxena, A.; Tavakol, M.; Fletcher, G.; and Pechenizkiy, M. 2022.
\newblock Survey on fair reinforcement learning: Theory and practice.
\newblock \emph{arXiv preprint arXiv:2205.10032}.

\bibitem[{Ge et~al.(2022)Ge, Zhao, Yu, Paul, Hu, Hsieh, and Zhang}]{ge2022toward}
Ge, Y.; Zhao, X.; Yu, L.; Paul, S.; Hu, D.; Hsieh, C.-C.; and Zhang, Y. 2022.
\newblock Toward Pareto efficient fairness-utility trade-off in recommendation through reinforcement learning.
\newblock In \emph{Proceedings of the fifteenth ACM international conference on web search and data mining}, 316--324.

\bibitem[{Ghodsi and Mirfakhar(2021)}]{ghodsi2021online}
Ghodsi, M.; and Mirfakhar, A. 2021.
\newblock Online Fair Revenue Maximizing Cake Division with Non-Contiguous Pieces in Adversarial Bandits.
\newblock \emph{arXiv preprint arXiv:2111.14387}.

\bibitem[{Gillen et~al.(2018)Gillen, Jung, Kearns, and Roth}]{gillen2018online}
Gillen, S.; Jung, C.; Kearns, M.; and Roth, A. 2018.
\newblock Online learning with an unknown fairness metric.
\newblock \emph{Advances in neural information processing systems}, 31.

\bibitem[{Grupen, Selman, and Lee(2022)}]{grupen2022cooperative}
Grupen, N.~A.; Selman, B.; and Lee, D.~D. 2022.
\newblock Cooperative multi-agent fairness and equivariant policies.
\newblock In \emph{Proceedings of the AAAI Conference on Artificial Intelligence}, volume 36 (9), 9350--9359.

\bibitem[{Hacker(2018)}]{hacker2018teaching}
Hacker, P. 2018.
\newblock Teaching fairness to artificial intelligence: Existing and novel strategies against algorithmic discrimination under EU law.
\newblock \emph{Common Market Law Review}, 55(4).

\bibitem[{Hao et~al.(2023)Hao, Xu, Zhang, Yang, and Muntean}]{hao2023computing}
Hao, H.; Xu, C.; Zhang, W.; Yang, S.; and Muntean, G.-M. 2023.
\newblock Computing Offloading with Fairness Guarantee: A Deep Reinforcement Learning Method.
\newblock \emph{IEEE Transactions on Circuits and Systems for Video Technology}.

\bibitem[{Henzinger et~al.(2023)Henzinger, Karimi, Kueffner, and Mallik}]{henzinger2023runtime}
Henzinger, T.; Karimi, M.; Kueffner, K.; and Mallik, K. 2023.
\newblock Runtime Monitoring of Dynamic Fairness Properties.
\newblock In \emph{Proceedings of the 2023 ACM Conference on Fairness, Accountability, and Transparency}, 604--614.

\bibitem[{Huang et~al.(2022)Huang, Labille, Wu, Lee, and Heffernan}]{huang2022contextual}
Huang, W.; Labille, K.; Wu, X.; Lee, D.; and Heffernan, N. 2022.
\newblock Achieving user-side fairness in contextual bandits.
\newblock \emph{Human-Centric Intelligent Systems}, 2(3-4): 81--94.

\bibitem[{Huang, Zhang, and Wu(2022)}]{huang2022causal}
Huang, W.; Zhang, L.; and Wu, X. 2022.
\newblock Achieving counterfactual fairness for causal bandit.
\newblock In \emph{Proceedings of the AAAI Conference on Artificial Intelligence}, volume 36(6), 6952--6959.

\bibitem[{Jabbari et~al.(2017)Jabbari, Joseph, Kearns, Morgenstern, and Roth}]{jabbari2017fairness}
Jabbari, S.; Joseph, M.; Kearns, M.; Morgenstern, J.; and Roth, A. 2017.
\newblock Fairness in reinforcement learning.
\newblock In \emph{International conference on machine learning}, 1617--1626. PMLR.

\bibitem[{Jiang and Lu(2019)}]{jiang2019learning}
Jiang, J.; and Lu, Z. 2019.
\newblock Learning fairness in multi-agent systems.
\newblock \emph{Advances in Neural Information Processing Systems}, 32.

\bibitem[{Kaufmann et~al.(2023)Kaufmann, Weng, Bengs, and H{\"u}llermeier}]{kaufmann2023survey}
Kaufmann, T.; Weng, P.; Bengs, V.; and H{\"u}llermeier, E. 2023.
\newblock A Survey of Reinforcement Learning from Human Feedback.
\newblock \emph{arXiv preprint arXiv:2312.14925}.

\bibitem[{Kushner and Whiting(2004)}]{kushner2004convergence}
Kushner, H.~J.; and Whiting, P.~A. 2004.
\newblock Convergence of proportional-fair sharing algorithms under general conditions.
\newblock \emph{IEEE transactions on wireless communications}, 3(4): 1250--1259.

\bibitem[{Lattimore and Szepesv{\'a}ri(2020)}]{lattimore2020bandit}
Lattimore, T.; and Szepesv{\'a}ri, C. 2020.
\newblock \emph{Bandit algorithms}.
\newblock Cambridge University Press.

\bibitem[{Li et~al.(2020)Li, Ma, Xia, Zhao, and Yang}]{li2020fairness}
Li, C.; Ma, X.; Xia, L.; Zhao, Q.; and Yang, J. 2020.
\newblock Fairness control of traffic light via deep reinforcement learning.
\newblock In \emph{2020 IEEE 16th International Conference on Automation Science and Engineering (CASE)}, 652--658. IEEE.

\bibitem[{Li, Liu, and Ji(2019)}]{li2019combinatorial}
Li, F.; Liu, J.; and Ji, B. 2019.
\newblock Combinatorial sleeping bandits with fairness constraints.
\newblock \emph{IEEE Transactions on Network Science and Engineering}, 7(3): 1799--1813.

\bibitem[{Li et~al.(2023)Li, Chen, Xu, Ge, Tan, Liu, and Zhang}]{li2023fairness}
Li, Y.; Chen, H.; Xu, S.; Ge, Y.; Tan, J.; Liu, S.; and Zhang, Y. 2023.
\newblock Fairness in Recommendation: Foundations, Methods, and Applications.
\newblock \emph{ACM Transactions on Intelligent Systems and Technology}, 14(5): 1--48.

\bibitem[{Li et~al.(2021)Li, Chen, Xu, Ge, and Zhang}]{li2021towards}
Li, Y.; Chen, H.; Xu, S.; Ge, Y.; and Zhang, Y. 2021.
\newblock Towards personalized fairness based on causal notion.
\newblock In \emph{Proceedings of the 44th International ACM SIGIR Conference on Research and Development in Information Retrieval}, 1054--1063.

\bibitem[{Liu et~al.(2018)Liu, Chen, Tang, Xu, and Piao}]{liu2018energy}
Liu, C.~H.; Chen, Z.; Tang, J.; Xu, J.; and Piao, C. 2018.
\newblock Energy-efficient UAV control for effective and fair communication coverage: A deep reinforcement learning approach.
\newblock \emph{IEEE Journal on Selected Areas in Communications}, 36(9): 2059--2070.

\bibitem[{Liu(2023)}]{liu2023transforming}
Liu, G. K.-M. 2023.
\newblock Transforming Human Interactions with AI via Reinforcement Learning with Human Feedback (RLHF).
\newblock \emph{Massachusetts Institute of Technology}.

\bibitem[{Liu et~al.(2020)Liu, Liu, Tang, Liao, Chen, and Heng}]{liu2020balancing}
Liu, W.; Liu, F.; Tang, R.; Liao, B.; Chen, G.; and Heng, P.~A. 2020.
\newblock Balancing between accuracy and fairness for interactive recommendation with reinforcement learning.
\newblock \emph{Advances in Knowledge Discovery and Data Mining}, 12084: 155.

\bibitem[{Liu et~al.(2017)Liu, Radanovic, Dimitrakakis, Mandal, and Parkes}]{liu2017calibrated}
Liu, Y.; Radanovic, G.; Dimitrakakis, C.; Mandal, D.; and Parkes, D.~C. 2017.
\newblock Calibrated fairness in bandits.
\newblock \emph{arXiv preprint arXiv:1707.01875}.

\bibitem[{Mandal and Gan(2022)}]{mandal2022socially}
Mandal, D.; and Gan, J. 2022.
\newblock Socially fair reinforcement learning.
\newblock \emph{arXiv preprint arXiv:2208.12584}.

\bibitem[{Mehrabi et~al.(2021)Mehrabi, Morstatter, Saxena, Lerman, and Galstyan}]{mehrabi2021survey}
Mehrabi, N.; Morstatter, F.; Saxena, N.; Lerman, K.; and Galstyan, A. 2021.
\newblock A survey on bias and fairness in machine learning.
\newblock \emph{ACM computing surveys (CSUR)}, 54(6): 1--35.

\bibitem[{Metevier et~al.(2019)Metevier, Giguere, Brockman, Kobren, Brun, Brunskill, and Thomas}]{metevier2019offline}
Metevier, B.; Giguere, S.; Brockman, S.; Kobren, A.; Brun, Y.; Brunskill, E.; and Thomas, P.~S. 2019.
\newblock Offline contextual bandits with high probability fairness guarantees.
\newblock \emph{Advances in neural information processing systems}, 32.

\bibitem[{Miconi(2017)}]{miconi2017impossibility}
Miconi, T. 2017.
\newblock The impossibility of" fairness": a generalized impossibility result for decisions.
\newblock \emph{arXiv preprint arXiv:1707.01195}.

\bibitem[{Nabi, Malinsky, and Shpitser(2019)}]{nabi2019learning}
Nabi, R.; Malinsky, D.; and Shpitser, I. 2019.
\newblock Learning optimal fair policies.
\newblock In \emph{International Conference on Machine Learning}, 4674--4682. PMLR.

\bibitem[{Patil et~al.(2021)Patil, Ghalme, Nair, and Narahari}]{patil2021achieving}
Patil, V.; Ghalme, G.; Nair, V.; and Narahari, Y. 2021.
\newblock Achieving fairness in the stochastic multi-armed bandit problem.
\newblock \emph{The Journal of Machine Learning Research}, 22(1): 7885--7915.

\bibitem[{Petrangeli et~al.(2014)Petrangeli, Claeys, Latr{\'e}, Famaey, and De~Turck}]{petrangeli2014multi}
Petrangeli, S.; Claeys, M.; Latr{\'e}, S.; Famaey, J.; and De~Turck, F. 2014.
\newblock A multi-agent Q-Learning-based framework for achieving fairness in HTTP Adaptive Streaming.
\newblock In \emph{2014 IEEE Network Operations and Management Symposium (NOMS)}, 1--9. IEEE.

\bibitem[{Pirotta, Parisi, and Restelli(2015)}]{pirotta2015multi}
Pirotta, M.; Parisi, S.; and Restelli, M. 2015.
\newblock Multi-objective reinforcement learning with continuous pareto frontier approximation.
\newblock In \emph{Proceedings of the AAAI conference on artificial intelligence}, volume 29 (1).

\bibitem[{Qi et~al.(2020)Qi, Hu, Huang, Wen, and Lu}]{qi2020uavcontrol}
Qi, H.; Hu, Z.; Huang, H.; Wen, X.; and Lu, Z. 2020.
\newblock Energy Efficient 3-D UAV Control for Persistent Communication Service and Fairness: A Deep Reinforcement Learning Approach.
\newblock \emph{IEEE Access}, 8: 53172--53184.

\bibitem[{Rabin(1993)}]{rabin1993incorporating}
Rabin, M. 1993.
\newblock Incorporating fairness into game theory and economics.
\newblock \emph{The American economic review}, 1281--1302.

\bibitem[{Raeis and Leon-Garcia(2021)}]{raeis2021trafficcontrol}
Raeis, M.; and Leon-Garcia, A. 2021.
\newblock A Deep Reinforcement Learning Approach for Fair Traffic Signal Control.
\newblock In \emph{2021 IEEE International Intelligent Transportation Systems Conference (ITSC)}, 2512--2518.

\bibitem[{Rawls(1971)}]{rawls1971atheory}
Rawls, J. 1971.
\newblock A theory of justice.
\newblock \emph{Cambridge (Mass.)}.

\bibitem[{Siddique, Weng, and Zimmer(2020)}]{siddique2020learning}
Siddique, U.; Weng, P.; and Zimmer, M. 2020.
\newblock Learning Fair Policies in Multi-Objective (Deep) Reinforcement Learning with Average and Discounted Rewards.
\newblock In \emph{International Conference on Machine Learning}, 8905--8915. PMLR.

\bibitem[{Singh et~al.(2020)Singh, Halpern, Thain, Christakopoulou, Chi, Chen, and Beutel}]{singh2020healthyrecommendation}
Singh, A.; Halpern, Y.; Thain, N.; Christakopoulou, K.; Chi, E.; Chen, J.; and Beutel, A. 2020.
\newblock Building healthy recommendation sequences for everyone: A safe reinforcement learning approach.
\newblock \emph{FAccTRec Workshop}.

\bibitem[{Skarlicki and Folger(1997)}]{skarlicki1997retaliation}
Skarlicki, D.~P.; and Folger, R. 1997.
\newblock Retaliation in the workplace: The roles of distributive, procedural, and interactional justice.
\newblock \emph{Journal of applied Psychology}, 82(3): 434.

\bibitem[{Steck, van Zwol, and Johnson(2015)}]{steck2015interactive}
Steck, H.; van Zwol, R.; and Johnson, C. 2015.
\newblock Interactive recommender systems: Tutorial.
\newblock In \emph{Proceedings of the 9th ACM Conference on Recommender Systems}, 359--360.

\bibitem[{Sutton and Barto(2018)}]{sutton2018reinforcement}
Sutton, R.~S.; and Barto, A.~G. 2018.
\newblock \emph{Reinforcement learning: An introduction}.
\newblock MIT press.

\bibitem[{Tong et~al.(2021)Tong, Chen, Zhou, and Sun}]{tong2021qoefairness}
Tong, L.; Chen, Y.; Zhou, X.; and Sun, Y. 2021.
\newblock Qoe-fairness tradeoff scheme for dynamic spectrum allocation based on deep reinforcement learning.
\newblock In \emph{Proceedings of the 5th International Conference on Computer Science and Application Engineering}, 1--7.

\bibitem[{Valkanis et~al.(2022)Valkanis, Papadimitriou, Beletsioti, Varvarigos, and Nicopolitidis}]{valkanis2022traffic}
Valkanis, A.; Papadimitriou, G.; Beletsioti, G.; Varvarigos, E.; and Nicopolitidis, P. 2022.
\newblock Efficiency and fairness improvement for elastic optical networks using reinforcement learning-based traffic prediction.
\newblock \emph{Journal of Optical Communications and Networking}, 14(3): 25--42.

\bibitem[{Vlasceanu and Amodio(2022)}]{vlasceanu2022propagation}
Vlasceanu, M.; and Amodio, D.~M. 2022.
\newblock Propagation of societal gender inequality by internet search algorithms.
\newblock \emph{Proceedings of the National Academy of Sciences}, 119(29): e2204529119.

\bibitem[{Wang et~al.(2021{\natexlab{a}})Wang, Bai, Sun, and Joachims}]{wang2021fairnessexposure}
Wang, L.; Bai, Y.; Sun, W.; and Joachims, T. 2021{\natexlab{a}}.
\newblock Fairness of Exposure in Stochastic Bandits.
\newblock In Meila, M.; and Zhang, T., eds., \emph{Proceedings of the 38th International Conference on Machine Learning}, volume 139 of \emph{Proceedings of Machine Learning Research}, 10686--10696. PMLR.

\bibitem[{Wang and Deng(2020)}]{wang2020facialrecognition}
Wang, M.; and Deng, W. 2020.
\newblock Mitigating Bias in Face Recognition Using Skewness-Aware Reinforcement Learning.
\newblock In \emph{Proceedings of the IEEE/CVF Conference on Computer Vision and Pattern Recognition (CVPR)}.

\bibitem[{Wang et~al.(2021{\natexlab{b}})Wang, Wang, Beutel, Prost, Chen, and Chi}]{wang2021tradeoff}
Wang, Y.; Wang, X.; Beutel, A.; Prost, F.; Chen, J.; and Chi, E.~H. 2021{\natexlab{b}}.
\newblock Understanding and Improving Fairness-Accuracy Trade-offs in Multi-Task Learning.
\newblock In \emph{Proceedings of the 27th ACM SIGKDD Conference on Knowledge Discovery \& Data Mining}, KDD '21, 1748–1757. New York, NY, USA: Association for Computing Machinery.
\newblock ISBN 9781450383325.

\bibitem[{Weng(2019)}]{weng2019fairness}
Weng, P. 2019.
\newblock Fairness in reinforcement learning.
\newblock \emph{arXiv preprint arXiv:1907.10323}.

\bibitem[{Wolff(1998)}]{wolff1998fairness}
Wolff, J. 1998.
\newblock Fairness, respect, and the egalitarian ethos.
\newblock \emph{Philosophy \& public affairs}, 27(2): 97--122.

\bibitem[{Yamazaki and Yamamoto(2021)}]{yamazaki2021congestion}
Yamazaki, M.; and Yamamoto, M. 2021.
\newblock Fairness Improvement of Congestion Control with Reinforcement Learning.
\newblock \emph{Journal of Information Processing 29}, 592--595.

\bibitem[{Zhang and Shah(2014)}]{zhang2014maximinfairness}
Zhang, C.; and Shah, J.~A. 2014.
\newblock Fairness in Multi-Agent Sequential Decision-Making.
\newblock In Ghahramani, Z.; Welling, M.; Cortes, C.; Lawrence, N.; and Weinberger, K., eds., \emph{Advances in Neural Information Processing Systems}, volume~27. Curran Associates, Inc.

\bibitem[{Zhang and Wang(2021)}]{zhang2021recommendationfairness}
Zhang, D.; and Wang, J. 2021.
\newblock Recommendation Fairness: From Static to Dynamic.
\newblock \emph{CoRR}, abs/2109.03150.

\bibitem[{Zhang and Liu(2021)}]{zhang2021survey}
Zhang, X.; and Liu, M. 2021.
\newblock Fairness in learning-based sequential decision algorithms: A survey.
\newblock \emph{Handbook of Reinforcement Learning and Control. Cham: Springer International Publishing}, 525--555.

\bibitem[{Zhong et~al.(2024)Zhong, Deng, Su, Wu, and Zhang}]{zhong2024provable}
Zhong, H.; Deng, Z.; Su, W.~J.; Wu, Z.~S.; and Zhang, L. 2024.
\newblock Provable Multi-Party Reinforcement Learning with Diverse Human Feedback.
\newblock \emph{arXiv preprint arXiv:2403.05006}.

\bibitem[{Zhu et~al.(2023)Zhu, Hu, Liang, Zhang, Jin, and Liu}]{zhu2023fairness}
Zhu, J.; Hu, M.; Liang, X.; Zhang, A.; Jin, R.; and Liu, R. 2023.
\newblock Fairness-Sensitive Policy-Gradient Reinforcement Learning for Reducing Bias in Robotic Assistance.
\newblock \emph{arXiv preprint arXiv:2306.04167}.

\bibitem[{Zhu and Oh(2018)}]{zhu2018multirobot}
Zhu, Q.; and Oh, J. 2018.
\newblock Deep Reinforcement Learning for Fairness in Distributed Robotic Multi-type Resource Allocation.
\newblock In \emph{2018 17th IEEE International Conference on Machine Learning and Applications (ICMLA)}, 460--466.

\bibitem[{Zimmer et~al.(2021)Zimmer, Glanois, Siddique, and Weng}]{zimmer2021learning}
Zimmer, M.; Glanois, C.; Siddique, U.; and Weng, P. 2021.
\newblock Learning fair policies in decentralized cooperative multi-agent reinforcement learning.
\newblock In \emph{International Conference on Machine Learning}, 12967--12978. PMLR.

\end{thebibliography}

\end{document}